\documentclass{llncs}
\usepackage{llncsdoc}

\usepackage[english]{babel}
\usepackage[utf8]{inputenc}
\usepackage{enumerate}
\usepackage{cite}
\usepackage{graphicx}  
\usepackage{subfig}
\usepackage{svg}
\usepackage{lscape}

\usepackage{tabularx}
\usepackage{colortbl}
\usepackage{hhline}
\usepackage{hyperref}

\usepackage{pgf}
\usepackage{tikz}
\usepackage[utf8]{inputenc}
\usetikzlibrary{arrows,automata}
\usetikzlibrary{positioning}
\usepackage{tabularx,ragged2e,booktabs,caption}

\tikzset{
	state/.style={
		rectangle,
		rounded corners,
		draw=black,
		minimum height=2em,
		inner sep=2pt,
		text centered,
	},
}

\begin{document}
%
\title{Predicting city safety perception based on visual image content}
\titlerunning{Predicting city safety perception based on visual image content}


\author{Sergio F. Acosta, Jorge E. Camargo}
\institute{Universidad Nacional de Colombia - UnSecureLab Research Group\\
\email{\{sfacostale,jecamargom\}@unal.edu.co}}

\authorrunning{Sergio Acosta, Jorge E. Camargo}

\maketitle

\begin{abstract}
Safety perception measurement has been a subject of interest in many cities of the world. This is due to its social relevance, and to its effect on some local economic activities. Even though people safety perception is a subjective topic, sometimes it is possible to find out common patterns given a restricted geographical and sociocultural context. This paper presents an approach that makes use of image processing and machine learning techniques to detect with high accuracy urban environment patterns that could affect citizen's safety perception. 
\end{abstract}


%


\section{Introduction}
Cities are spatial structures of a significant size. In the case of Bogotá-Colombia, the city has an urban area of $307,36 km^{2}$ , and it is divided into 20 well defined zones that are named localities.  This attribute makes it difficult to appreciate and experience them completely in just one round. This is the reason why the individual perception of a city is the result of a mixture of own experiences and experiences from others. Neighborhoods often differ in their demographics, such as income, and ethnicity of people that inhabits them, but also on how safe they are perceived \cite{Kominers2015}. Some of the recent works about automatic urban perception prediction have been based on Convolutional Neural Networks (CNN) \cite{Krizhevsky2012}, sets of between 100,000 and 1,000,000 images and with a wider territory scale approach. This work presents an approach based on a restricted geographical and sociocultural context, a modest image set, and on a technique called transfer learning. According to \cite{Gomez2016}, perception of security can be defined as:

\begin{quote}
Perception of security (PoS) refers to the subjective assessment of the risk and the magnitude of its consequences. The risk can be defined as the likelihood that an individual will experience the effect of danger, threats, or any adverse events.
\end{quote}



Projects such as \cite{Dubey2016a},\cite{Porzi2015},\cite{Ordonez2014} and \cite{Naik} have been focused on how to structure computational models that make it possible the automatic characterization of cities. These works have based their research on the visual appearance of city streets, and its association with the people perception. Misclassification is mainly caused by the huge variability in the set of images associated to the same group or tag.

In \cite{Donahue2013} it is proposed that the activation output of inner layer of a CNN trained for some other classification task can be used as a visual feature of an image in a different classification task. Based on this idea, the first urban perception model based CNN was proposed in \cite{Porzi2015}. CNN approach is also implemented in \cite{Dubey2016a}, where a CNN architecture is implemented in order to build a worldwide urban perception model.

The methodology described in \cite{Salesses2013} involves an image score estimation using the fraction of times this is selected over another image, then this is corrected by the ‘‘win’’ and ‘‘loss’’ ratios of all images with which it was compared during a visual survey. In \cite{Naik}, people perception obtained through visual surveys is converted to a ranked score for each image using the Microsoft Trueskill algorithm \cite{Herbrich2006}. In \cite{Dubey2016a} and \cite{Porzi2015} it is proposed to predict pairwise comparisons by training a CNN model directly from image pairs and their crowd-sourced comparisons, which is used to generate “synthetic” comparisons by taking random image pairs as input.

Our paper presents an alternative approach in the construction of a computational model whose purpose is to predict how safe may be a given Bogotá city zone. This has been carried out in a restricted sociocultural and geographical context. In order to restrict sociocultural context, just people living in Bogotá was invited to take the visual survey. Safety perception like humor can depend on a particular sociocultural context. This is why sometimes a joke that is fun in the USA may not be fun in Germany. Geographical context restriction means that just local street images were used. One motivation for not using foreign street images is that many of the urban environment found in other countries, e.g. Washington or Germany, does not exist in Bogotá. It is expected that these restrictions reduce the variability of the underling distribution as well as the noisy of data. The presented approach makes use of a technique called transfer learning. This takes a piece of a model that has already been trained on a related task and reusing it in a new model. In particular VGG19 \cite{DBLP:journals/corr/SimonyanZ14a} model, loaded with weights trained on ImageNet, has been used for generic image feature extraction. A particular city zone of $40 km^{2}$  was chosen for this work, and local people were asked to participate in the visual survey. Since Bogotá has an extremely heterogeneous urban environment, this restriction still guarantees a significant image variability.  As a result, a model with an accuracy of 81\%  was obtained, and it was used to predict a safety perception score for two other neighboring localities. In order to detect different patterns, prediction on neighboring localities was performed using their particular image sets.

\section{Materials and methods}
\subsection{Dataset Construction}
\subsubsection{Street Image Crawler}
A city street image crawler was built using the Google Street View API V3.0 along with a file that contained geographical limits of the target zones.

\subsubsection{Image filtering}
Since some collected images have no a wide view of a street, but a close-up of a building facade, or a totally black image, it was required to remove these images from the collected set. This was done by extracting SIFT local descriptors from each image, and excluding images with a descriptor count less than 420.
  
\subsubsection{Visual survey}
Visual surveys was carried out through a public web site. On this, users were shown two geo\-tagged street city images, and asked to click on one in response to the questions ‘‘which place looks safer?’’. In this way a bit of citizenship perception is obtained. An image pair and its associated comparison are the unit of information that will be used during the training task. The visual survey tool is online in \textbf{\textit{http://wmodi.com/}}. Figure \ref{fig:wmodi} shows this site.
\begin{center}
	\centering
	\includegraphics[width=0.65\linewidth,height=110pt]{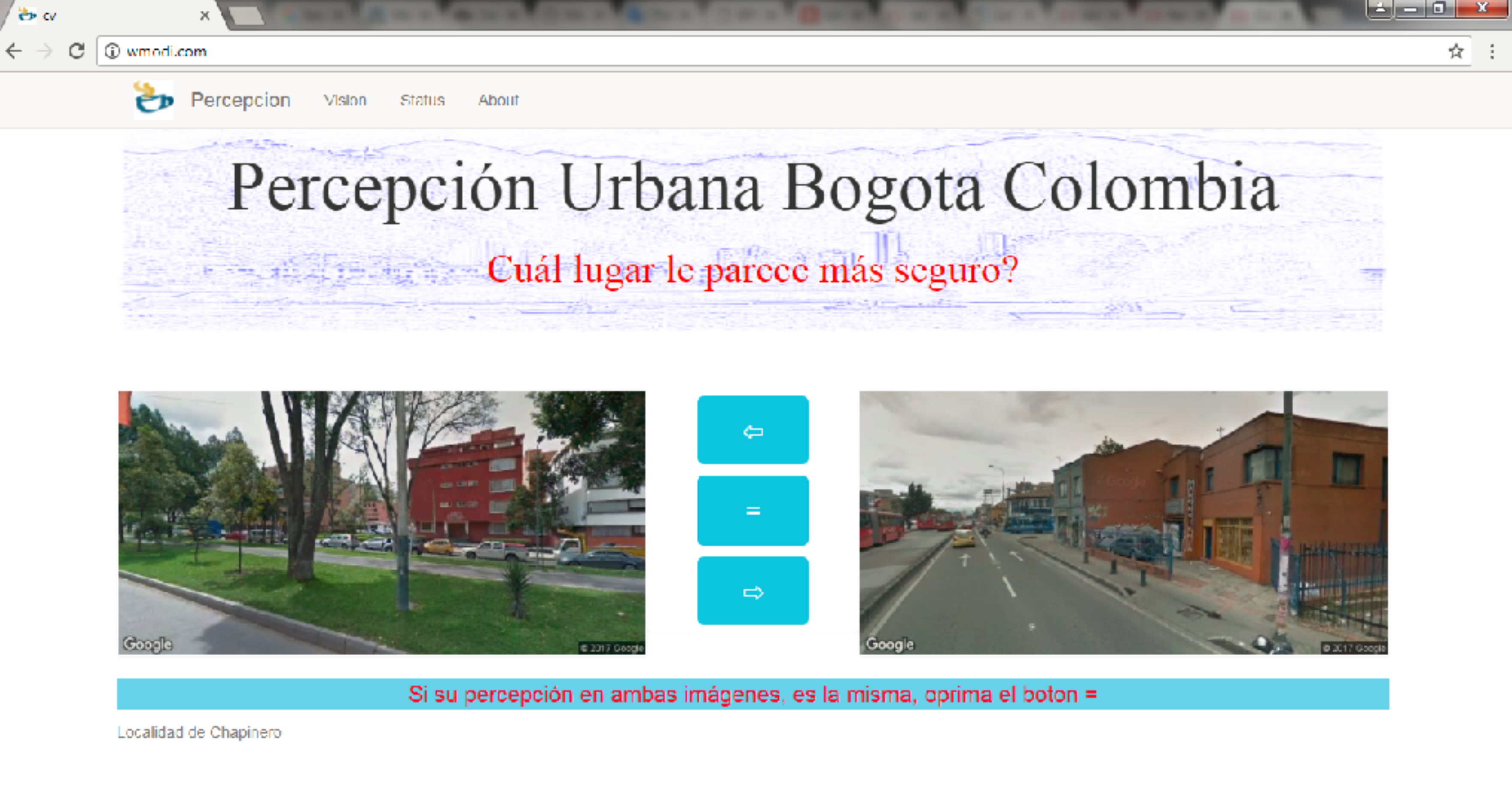}
	\captionof{figure}{Visual Survey Site}
	\label{fig:wmodi}
\end{center}

\subsubsection{Image serving policies}
Image pair serving policy and database housekeeping try to fulfill the following requirements. 
\begin{itemize}
	\item	Each image should have the same vote share.
	\item	Repeated image pair comparisons should be reduced to a single or no vote.
	\item	Same image comparison is not allowed and should be removed.
	\item	Comparisons of near images should be removed.
\end{itemize}

\subsubsection{Vote Coding and Database Structures} 
In each visual survey session, two street images were presented to the user. The user was asked to click on one image or on the equal button in order to answer the question: Which place looks safer?. Once the user click on an image or a button, the vote is coded as follows: If click on image 0 (left side image), vote is coded as 1; if click on button $equal$ vote is coded as 0. Finally, if click on image 1 (right side image), vote is coded as 2. For each visual survey session, the resulting vote code along with the involved left and right image identifiers were stored. There is also a database structure associated to each published image. This holds the image identifier, a positive perception counter, a negative perception counter and a neutral perception counter. If user click on image 0 (left side image), this image positive perception counter is increased by 1 and image 1 (right side image) negative perception counter is increased by 1. Same logic applies if user click on image 1 (right side image). Finally, if user click on button $equal$ both images neutral perception counter is increased by 1. This counters are used for each image basic perception percentage estimation.

\subsubsection{Collected Data}
5,505 images of the target zone were published. In one year, 17,703 image pair votes were collected. Each image participated in 6 $\pm 1$ image pair comparison session. The collected vote distribution based on its code is as follow: 5,657 code 0 votes, 5,946 code 1 votes, and 6,100 code 2 votes.

\subsection{Image feature extraction} 
IN THE VGG19 Keras\footnote{https://keras.io/} model it was removed its  fully-connected layer at the top of the network and loaded with the weights trained on ImageNet. Then, this model was used to obtain a row vector representation of each published image. This is a 512 values vector, i.e, the model is used like an image feature extractor and no training is required for this task. It is important to note that the goal of this research is to train a model that will be able to predict a visual survey vote based on a vector representation of a pair of city street images. That is, each training item is composed of two image feature vectors and their associated vote code.

\subsection{Training set construction} 
The construction of the training set involved the transformation of the collected visual survey votes into number vectors and its associated label (vote code). This task starts with the elaboration of a list of all the votes found in the database. From this list, image identifiers are replaced by their associated vector descriptors. Finally, images descriptors associated to each single vote are concatenated and annotated with the respective \{1, 2\}. Ties (code 0 votes) were not used, but just charged votes were used. If an actual vote has been annotated with vote code 1, it means that left image was better perceived than the right image. This means also that if image positions are exchanged, the resulting vote should be annotated with vote code 2.  This fact has been used to double the initial data set size. The resulting set of 24,092 (5,946*2 + 6,100*2) votes was split into training, validation and testing sets. A distribution of 65-7-28 was used. In this way 15,636 votes were used for training, 1,754 for validation and 6,702 votes for testing. Each descriptor vector corresponding to each image of the VGG19 was normalized applying the following expression,
\begin{center}
$ f_{i,j}=(f_{i,j}-\mu_{i})/\sigma_{i}$,
\end{center}
where $f_{i,j}$ is the $j$-th component of the $i$-th vector, ($\mu$) the mean, and ($\sigma$) the standard deviation. Each feature $i$ of each row vector $j$ is normalized by subtracting the associated mean and scaled by the associated standard deviation. Same normalization scheme was used on the image set of the neighboring localities, the model was used to predict on.

\subsubsection{Training phase}
Training sessions were performed using the TensorFlow\footnote{https://www.tensorflow.org/} machine learning framework. The implemented neural network configuration is shown in Figure \ref{fig:fcnn}

\begin{center}
	\centering
	\includegraphics[width=0.6\linewidth,height=50pt]{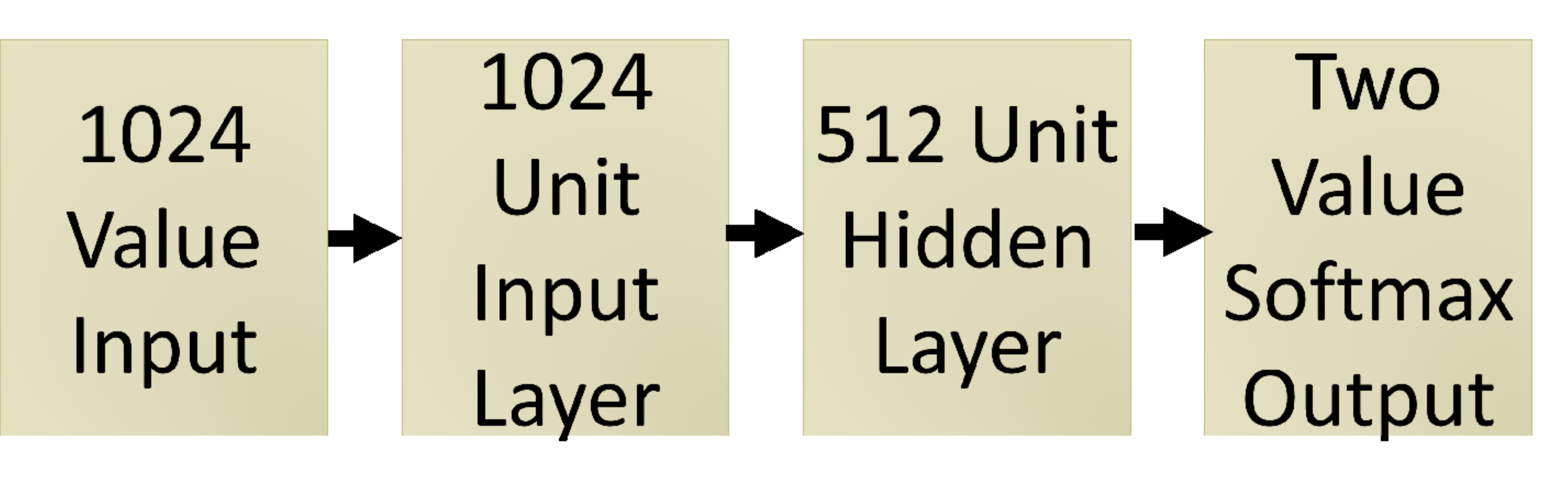}
	\captionof{figure}{Full Connected Neural Network Setup}
	\label{fig:fcnn}
\end{center}

The input data is the concatenation of image 0 and 1 VGG19 based on a 512 vector descriptor. This is a 1024 bin vector. A dropout technique was used as regularization method. Dropout rates of 0.5, 0.45 and 0.3 were applied to the input, hidden and output layers, respectively. The learning rate was set to 0.00001, and a mini-batch size was defined as 64. The AdamOptimizer \cite{DBLP:journals/corr/KingmaB14} method was used along with Cross Entropy Loss. Figure \ref{fig:nn4usp} shows training session cost curves.

\begin{center}
	\centering
	\includegraphics[width=0.6\linewidth,height=70pt]{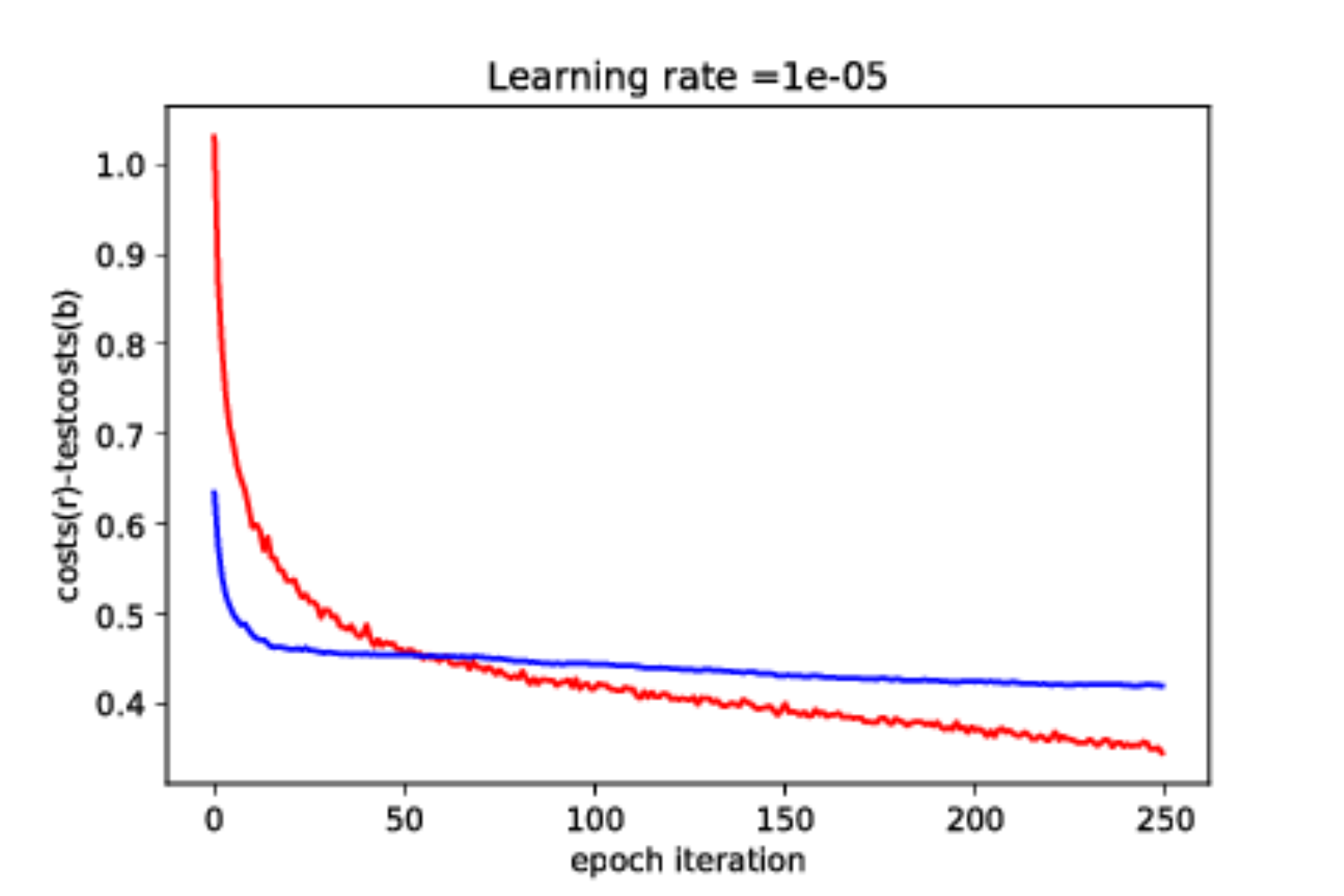}
	\captionof{figure}{Training session cost curves}
	\label{fig:nn4usp}
\end{center}

\subsubsection{Accuracy report}
Table \ref{tab:tstcnftrx} presents the confusion matrix for the testing set, for which an accuracy of 81.5\% was achieved.

\begin{center}
	\captionof{table}{Testing Confusion Matrix} \label{tab:tstcnftrx} 
	\begin{tabular}{l|l|c|c|c}
		\multicolumn{2}{c}{}&\multicolumn{2}{c}{Predicted Vote Code}&\\
		\cline{3-4}
		\multicolumn{2}{c|}{}&1&2&\multicolumn{1}{c}{Total}\\
		\cline{2-4}
		& 1 & 2,777 & 574& 3,351\\
		\cline{2-4}
		& 2 & 663 & 2,688 & 3,351\\
		\cline{2-4}
		\multicolumn{1}{c}{} & \multicolumn{1}{c}{Total} & \multicolumn{1}{c}{3,440} & \multicolumn{    1}{c}{3,262} & \multicolumn{1}{c}{6,702}\\
	\end{tabular}
	
\end{center}

\subsubsection{Synthetic vote generation}
Synthetic votes were generated dividing each locality collected image set into two same size groups. Each group holds images homogeneously distributed over the target locality area. Each group was divided into 10 subgroups. Then, each image from one group was paired with a randomly (uniform) selected image from each subgroup in the other group.

\subsubsection{Synthetic vote prediction}
In order to be able to make a map based on synthetic votes, the initial dummy vote label must be substituted by one based on the model prediction. The output of the softmax layer was used for this purpose. At first, the option with the higher probability was used to annotate the associated vote as 1 or 2. However, the absolute difference between the two probabilities must be higher than 0.25, otherwise the vote was annotated with the code 0. At the same time each image positive, negative and neutral perception counters were updated.

\subsubsection{Image Score Based on Perception Counters}
In section "Vote Coding and Database Structures" it was mentioned that published images had an structure associated to them and how its fields are updated during a visual survey session. This holds the image identifier, a positive perception counter, a negative perception counter and a neutral perception counter. Every image used in the synthetic vote prediction task has the same structure, and its perception counters are updated in the same way as during a visual survey session. If an image neutral perception counter is greater than 0, this value is redistributed between this image positive and negative counters. This redistribution is done by factors that are worked out form the perception counter summation of all images, which this image tied with. Finally, each image counter summation is normalized [0,1] and each counter turned into positive, negative and neutral safety perception percentages.

\section{Results}
\subsection{Actual and synthetic vote maps of same zone}
For the initial target zone training images were obtained, and a perception maps was built based on both, actual and synthetic votes. This was done in order to verify that colors patterns found in actual vote perception map are present in the synthetic vote perception map. For further exploration, actual\footnote{\href{http://wmodi.com/chapinero\_17703actualvote\_jun04\_2018\_imgscore}{http://wmodi.com/chapinero\_17703actualvote\_jun04\_2018\_imgscore}} and synthetic\footnote{\href{http://wmodi.com/chapinero\_55040NNsyntheticvote\_jun04\_2018\_imgscore}{http://wmodi.com/chapinero\_55040NNsyntheticvote\_jun04\_2018\_imgscore}} vote perception map are available in the project web site.

Figure \ref{fig:gradintcolor} shows the color gradient used on the map set. Here left green end indicates a 100\% percent of positive safety perception, and right red end 0\%. 

\begin{center}
	\centering
	\includegraphics[width=0.2\linewidth,height=10pt]{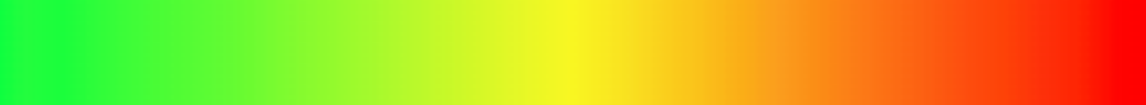}
	\captionof{figure}{Color gradient reference for image safety perception score}
	\label{fig:gradintcolor}
\end{center}

\subsection{Safety perception score prediction for other localities}
Model was used to generate synthetic votes on image pairs of different city zones. These image maps are available on line at left \footnote{\href{http://wmodi.com/usaquen\_94780NNsyntheticvote\_jun04\_2018\_imgscore}{http://wmodi.com/usaquen\_94780NNsyntheticvote\_jun04\_2018\_imgscore}} and right \footnote{\href{http://wmodi.com/martires\_37880NNsyntheticvote\_jun04\_2018\_imgscore}{http://wmodi.com/martires\_37880NNsyntheticvote\_jun04\_2018\_imgscore}} image map links.

\begin{center}
	\centering
	\includegraphics[width=0.2\linewidth,height=70pt]{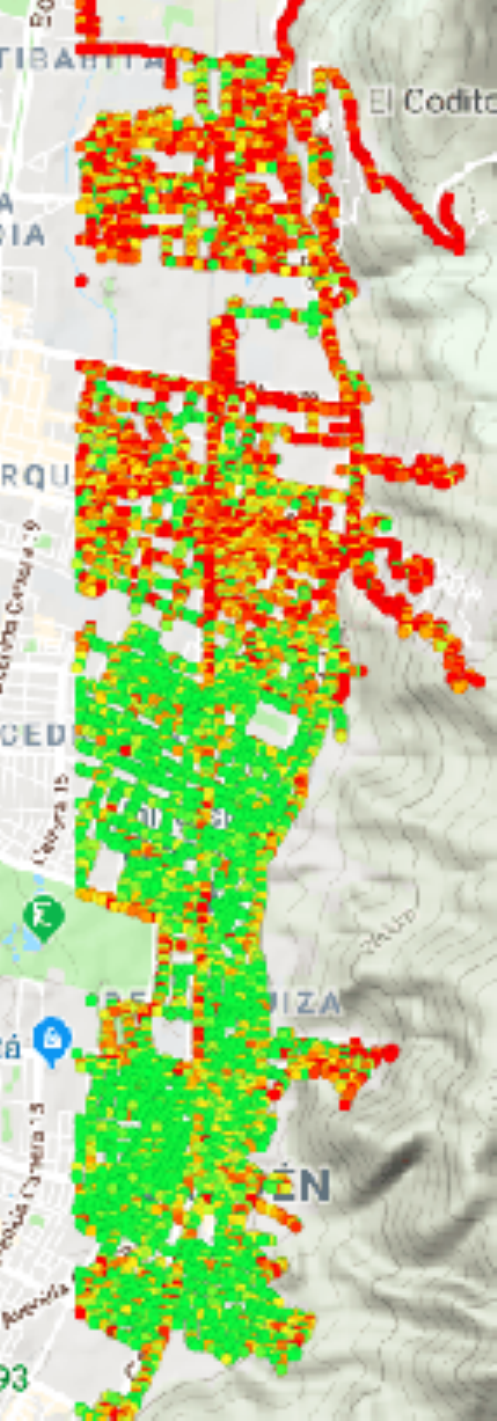}\hspace{1.2mm}%
    \includegraphics[width=0.35\linewidth,height=70pt]{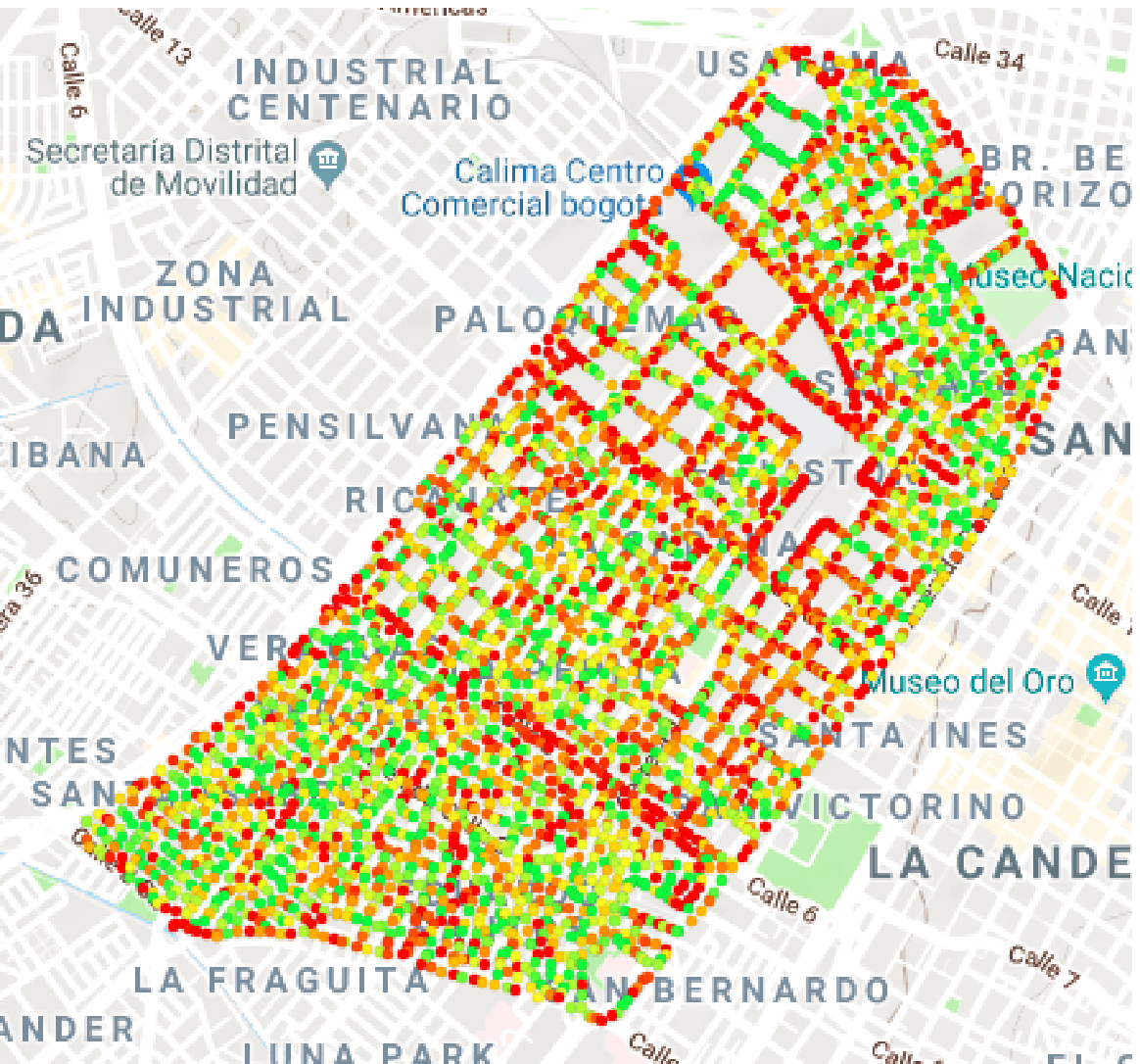}%
	\captionof{figure}{Left: $Usaquen$ zone 94,780 synthetic votes safety perception score map. Right: $Martires$ zone 37,880 synthetic votes safety perception score map}
	\label{fig:zone_predictions}
\end{center}

\subsection{Predicted image set}
Figures \ref{fig:zone_martirespre} and \ref{fig:zone_usaquenpre} are samples of other zone images whose score has been predicted by the system based on synthetic votes. It is worth noting that the trained model predicts with high precision the perception of test images. For instance, the predicted vote of left image in Figure \ref{fig:zone_martirespre} captures negative perception safety characteristics such as dirty houses, lonely streets, trash, etc., whilst the predicted vote of image in right image in Figure \label{fig:zone_usaquenpre} captures positive perception safety characteristics such illumination, green zones, clean streets, etc.

\begin{center}
	\centering
	\includegraphics[width=0.3\linewidth,height=78pt]{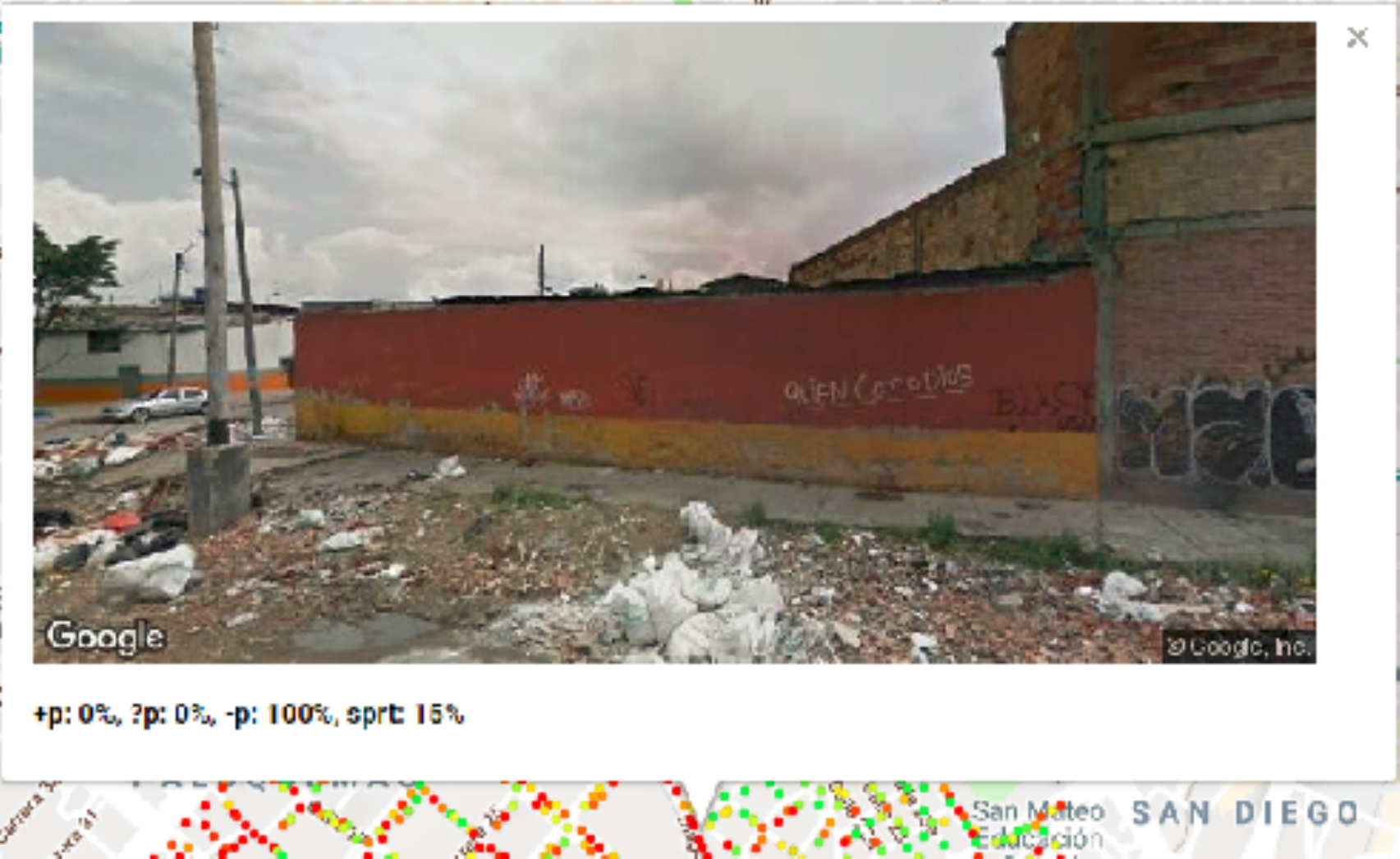}\hspace{1mm}%
    \includegraphics[width=0.3\linewidth,height=78pt]{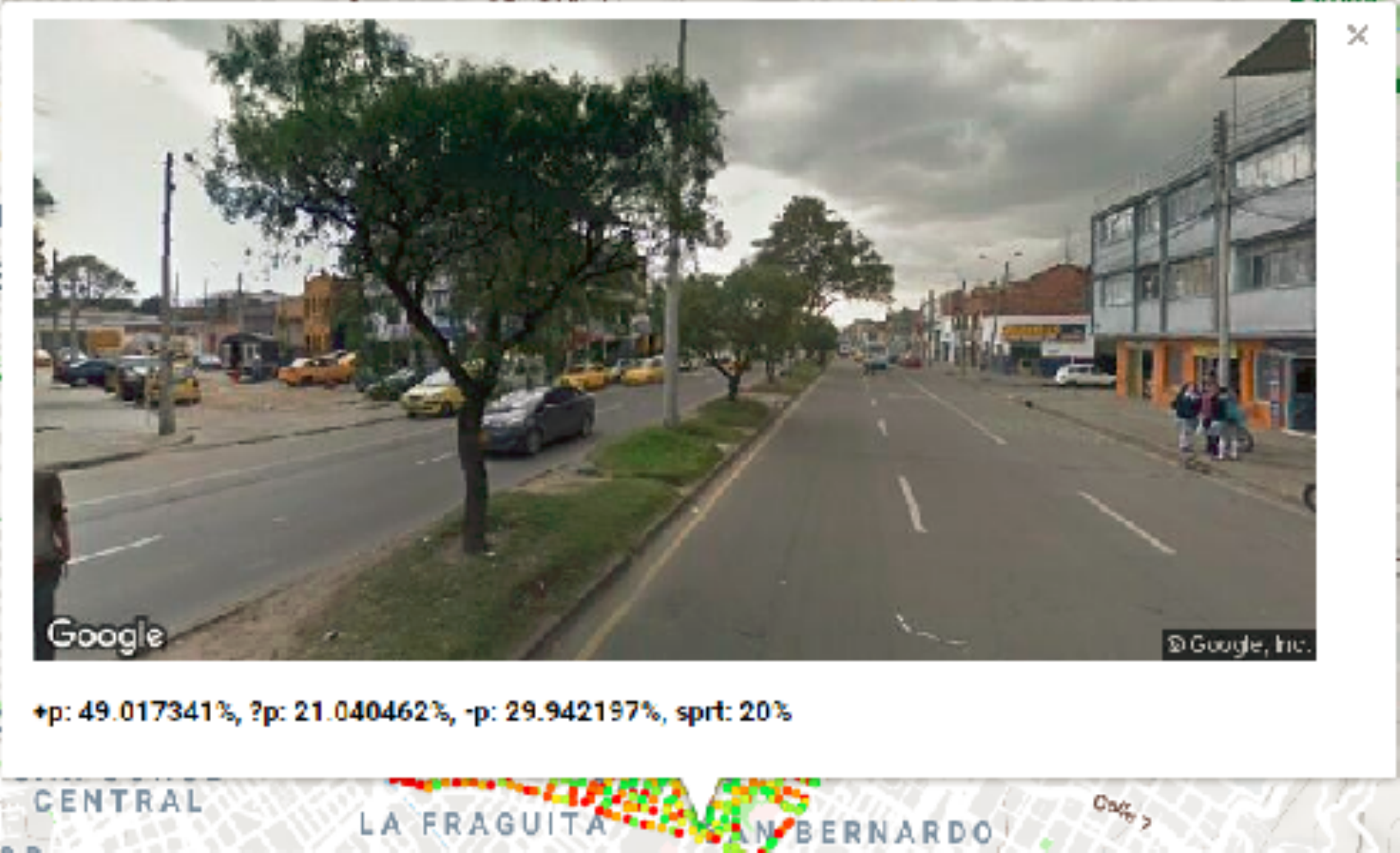}\hspace{1mm}%
    \includegraphics[width=0.3\linewidth,height=78pt]{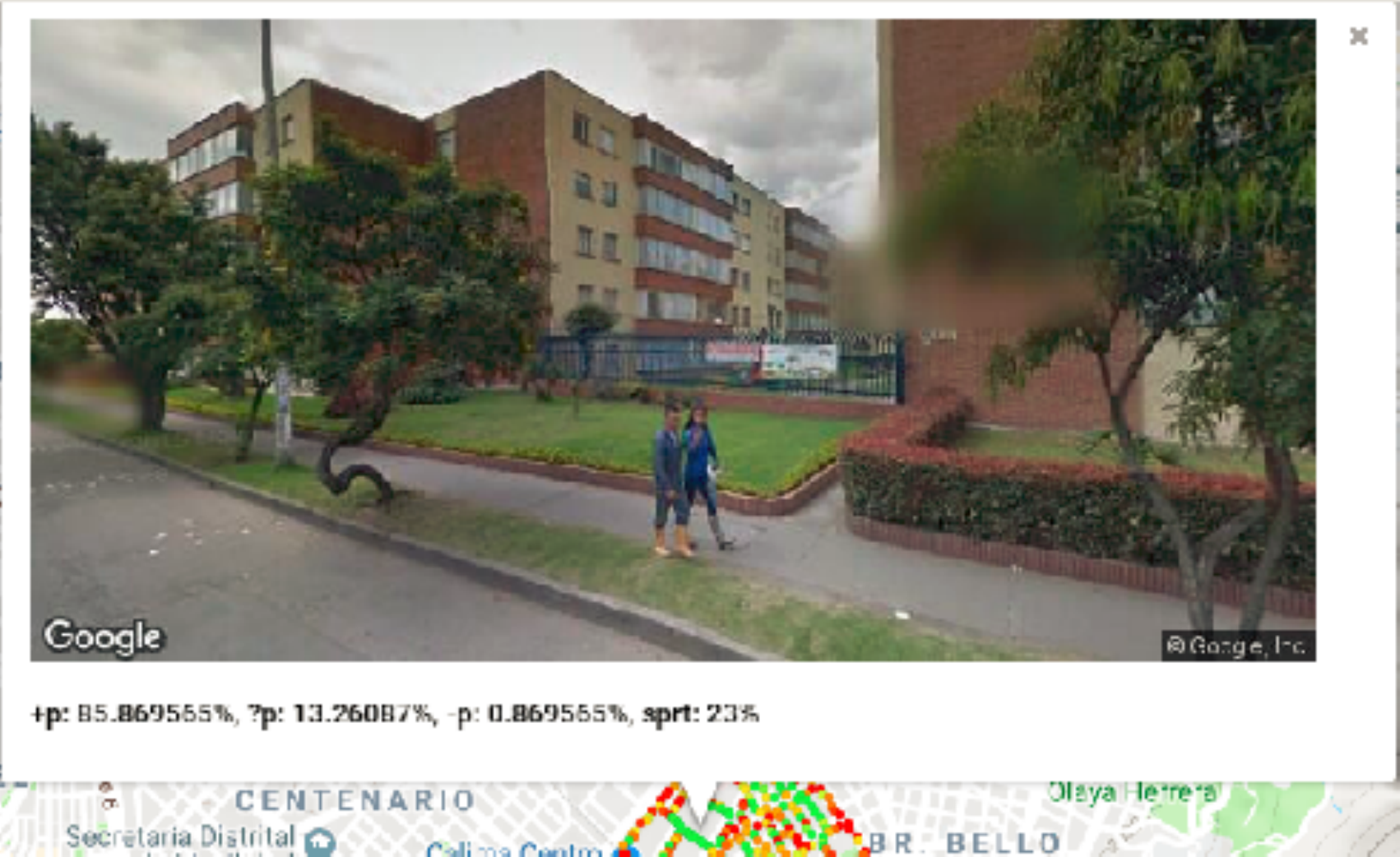}%
	\captionof{figure}{$Martires$ zone:0\%,49\%,86\% positive safety perception images }
	\label{fig:zone_martirespre}
\end{center}

\begin{center}
	\centering
	\includegraphics[width=0.3\linewidth,height=78pt]{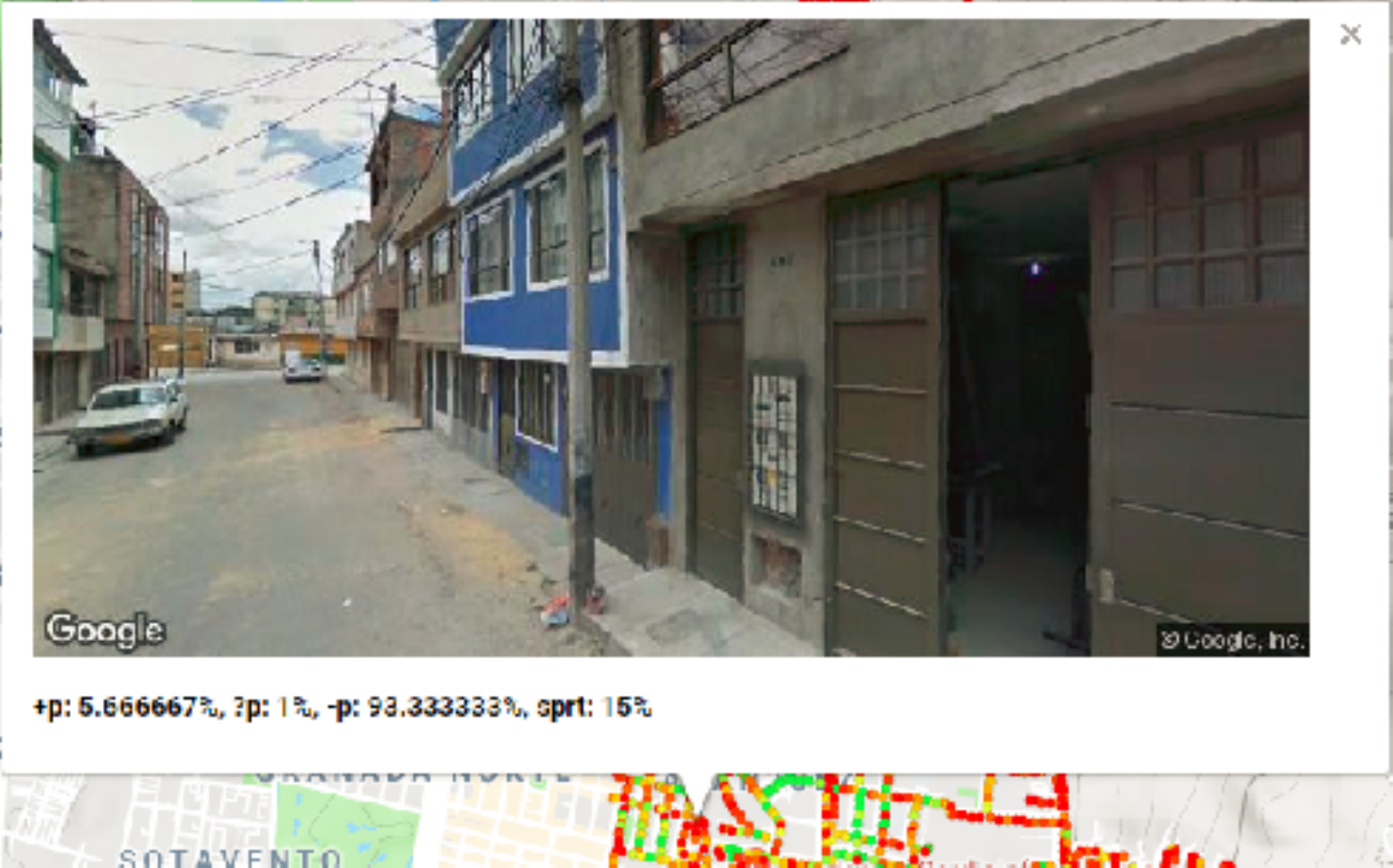}\hspace{1mm}%
    \includegraphics[width=0.3\linewidth,height=78pt]{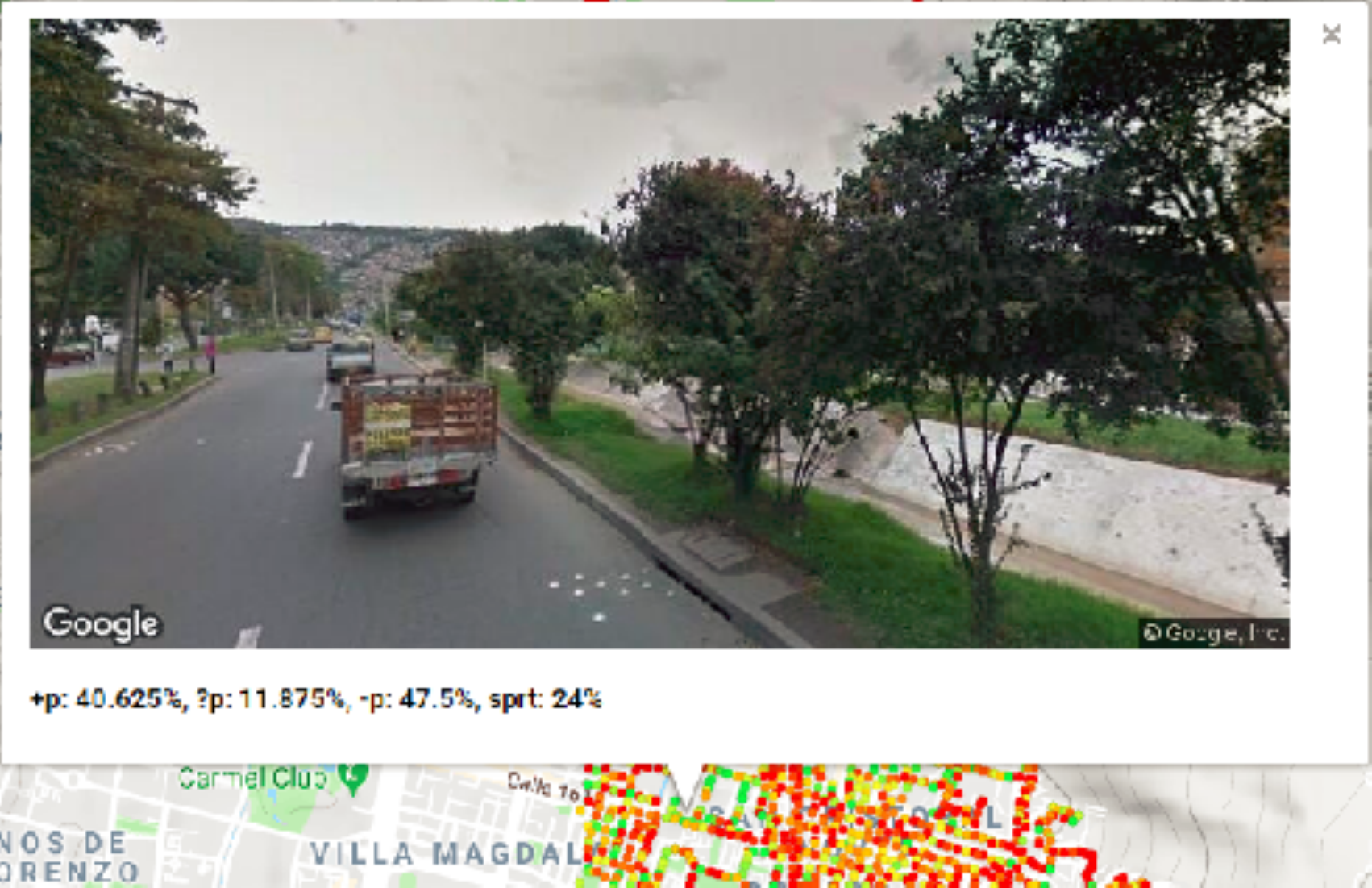}\hspace{1mm}%
    \includegraphics[width=0.3\linewidth,height=78pt]{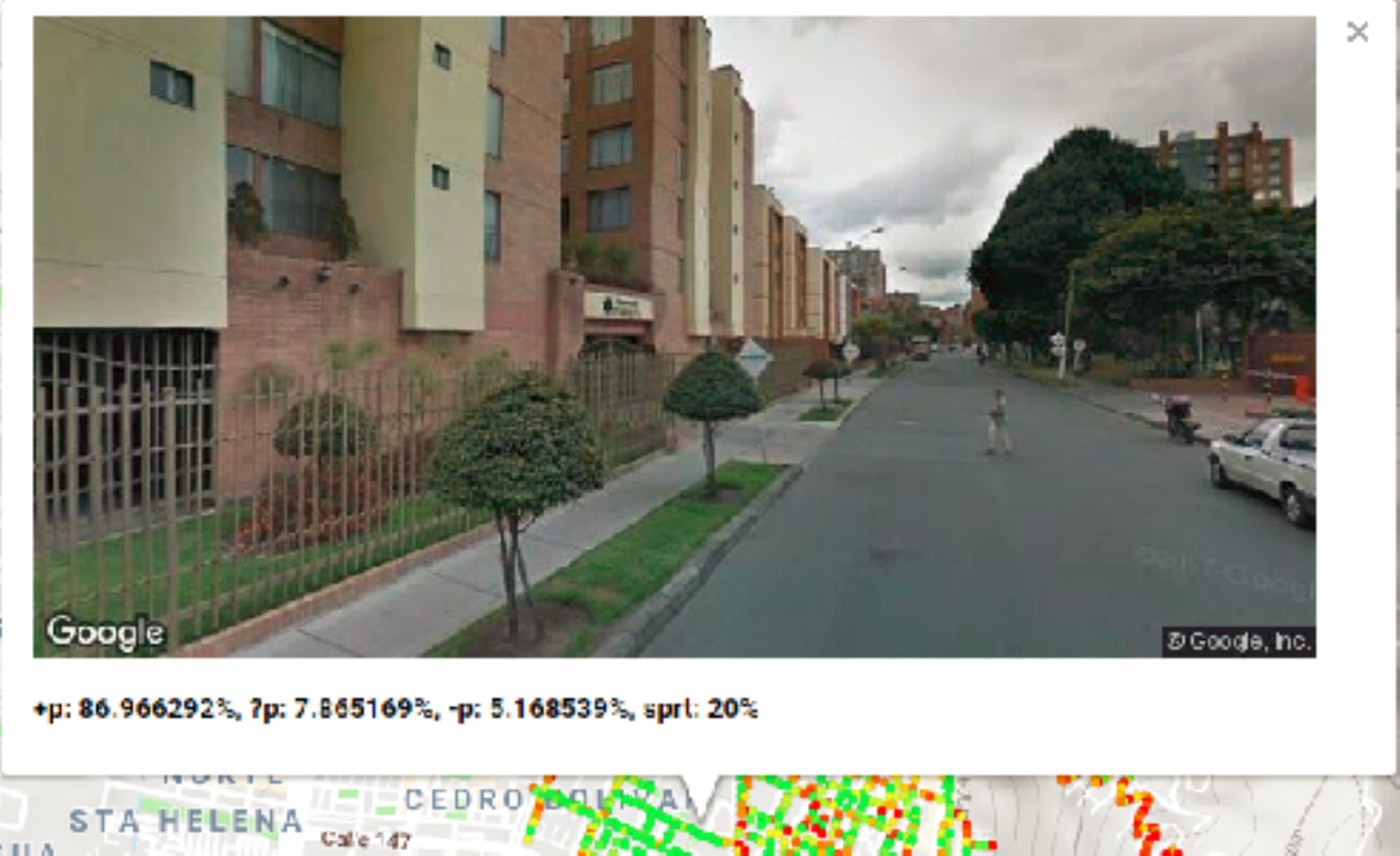}%
	\captionof{figure}{$Usaquen$ zone:6\%,41\%,87\% positive safety perception images }
	\label{fig:zone_usaquenpre}
\end{center}

\section{Conclusion and future work}
This paper presented a model that allows to predict citizen's safety perception using visual information of street images. The obtained results show a prediction accuracy of 81\%, which is higher than results obtained in recent state of the art methods. Up to our knowledge, this is the first time that this analysis is performed to Bogotá city. The presented method does not require a high computing capacity, that is, 1 model iterations per hour can be performed. This feature makes this approach appropriate for the development of an online tool. It is expected to carry out more exhaustive evaluations in order determine the robustness of the predictions as well as statistical stability. All these results and the results of an earlier model evaluated by us based on SVM with a smaller amount of votes are available at \textbf{\textit{http://wmodi.com/}}

\bibliographystyle{plain}       

\bibliography{template}

\begin{thebibliography}{10}

\bibitem{Donahue2013}
Jeff Donahue, Yangqing Jia, Oriol Vinyals, Judy Hoffman, Ning Zhang, Eric
  Tzeng, Trevor Darrell, Trevor Eecs, and Berkeley Edu.
\newblock {DeCAF: A Deep Convolutional Activation Feature for Generic Visual
  Recognition}.
\newblock 2013.

\bibitem{Dubey2016a}
Abhimanyu Dubey, Nikhil Naik, Devi Parikh, Ramesh Raskar, and C{\'{e}}sar~A.
  Hidalgo.
\newblock {Deep Learning the City : Quantifying Urban Perception At A Global
  Scale}.
\newblock pages 1--23, 2016.

\bibitem{Gomez2016}
Francisco G{\'{o}}mez, Andres Torres, Juan Galvis, Jorge Camargo, and Oscar
  Mart{\'{i}}nez.
\newblock {Hotspot mapping for perception of security}.
\newblock {\em IEEE 2nd International Smart Cities Conference: Improving the
  Citizens Quality of Life, ISC2 2016 - Proceedings}, pages 0--5, 2016.

\bibitem{Herbrich2006}
Ralf Herbrich, Tom Minka, and Thore Graepel.
\newblock {TrueSkill: A Bayesian Skill Rating System}.
\newblock {\em Advances in Neural Information Processing Systems}, 20:569--576,
  2006.

\bibitem{DBLP:journals/corr/KingmaB14}
Diederik~P. Kingma and Jimmy Ba.
\newblock Adam: {A} method for stochastic optimization.
\newblock {\em CoRR}, abs/1412.6980, 2014.

\bibitem{Kominers2015}
Scott~Duke Kominers, Edward~L Glaeser, C{\'{e}}sar~A Hidalgo, James Evans, Jay
  Garlapati, Lars Hansen, James Heckman, John~Eric Humphries, Jackie Hwang,
  Robert Sampson, Zak Stone, Erik Strand, Nina Tobio, and Mia Petkova.
\newblock {Do People Shape Cities, or Do Cities Shape People? The Co-evolution
  of Physical, Social, and Economic Change in Five Major U.S. Cities}.
\newblock 2015.

\bibitem{Krizhevsky2012}
Alex Krizhevsky, Ilya Sutskever, and Geoffrey~E Hinton.
\newblock {ImageNet Classification with Deep Convolutional Neural Networks}.
\newblock {\em Advances In Neural Information Processing Systems}, pages 1--9,
  2012.

\bibitem{Naik}
Nikhil Naik, Jade Philipoom, and Ramesh Raskar.
\newblock {Streetscore - Predicting the Perceived Safety of One Million
  Streetscapes}.

\bibitem{Ordonez2014}
Vicente Ordonez and Tamara~L. Berg.
\newblock {Learning high-level judgments of urban perception}.
\newblock {\em Lecture Notes in Computer Science (including subseries Lecture
  Notes in Artificial Intelligence and Lecture Notes in Bioinformatics)}, 8694
  LNCS(PART 6):494--510, 2014.

\bibitem{Porzi2015}
Lorenzo Porzi, Samuel~Rota Bul{\'{o}}, Bruno Lepri, and Elisa Ricci.
\newblock {Predicting and Understanding Urban Perception with Convolutional
  Neural Networks}.
\newblock pages 139--148, 2015.

\bibitem{Salesses2013}
Philip Salesses, Katja Schechtner, and C{\'{e}}sar~A. Hidalgo.
\newblock {The Collaborative Image of The City: Mapping the Inequality of Urban
  Perception}.
\newblock {\em PLoS ONE}, 2013.

\bibitem{DBLP:journals/corr/SimonyanZ14a}
Karen Simonyan and Andrew Zisserman.
\newblock Very deep convolutional networks for large-scale image recognition.
\newblock {\em CoRR}, abs/1409.1556, 2014.

\end{thebibliography}

\end{document}